\documentclass[letterpaper]{article} 
\usepackage{aaai2026}  
\usepackage{times}  
\usepackage{helvet}  
\usepackage{courier}  
\usepackage[hyphens]{url}  
\usepackage{graphicx} 
\usepackage{natbib}  
\usepackage{caption} 
\usepackage{amsfonts}
\usepackage{amsmath}
\usepackage{algorithm}
\usepackage{algorithmic}
\usepackage{xcolor} 
\usepackage{tabularray}
\UseTblrLibrary{booktabs} 

\usepackage{multirow}
\usepackage{newfloat}
\usepackage{listings}
\usepackage{colortbl}
\usepackage{ragged2e}

\DeclareCaptionStyle{ruled}{labelfont=normalfont,labelsep=colon,strut=off} 
\floatstyle{ruled}
\newfloat{listing}{tb}{lst}{}
\floatname{listing}{Listing}
\title{Empowering Semantic-Sensitive Underwater Image Enhancement with VLM}
\author {
   Guodong Fan\textsuperscript{\rm 1},
    Shengning Zhou\textsuperscript{\rm 1},
    Genji Yuan\textsuperscript{\rm 1},
    Huiyu Li\textsuperscript{\rm 2},
    Jingchun Zhou\textsuperscript{\rm 3},
    Jinjiang Li\textsuperscript{\rm 1} \thanks{Corresponding author.}
}
\affiliations {
    \textsuperscript{\rm 1}Shandong Technology and Business University, Yantai, China\\
    \textsuperscript{\rm 2}Shandong University of Finance and Economics, Jinan, China\\
    \textsuperscript{\rm 3}Dalian Maritime University, Dalian, China\\
    fgd96@outlook.com, 2023420043@sdtbu.edu.cn, 202414175@sdtbu.edu.cn, \\ huiyuroy@163.com, zhoujingchun03@qq.com, lijinjiang@gmail.com
}

\usepackage{bibentry}

\begin{document}

\maketitle

\begin{abstract}
In recent years, learning-based underwater image enhancement (UIE) techniques have rapidly evolved. However, distribution shifts between high-quality enhanced outputs and natural images can hinder semantic cue extraction for downstream vision tasks, thereby limiting the adaptability of existing enhancement models. To address this challenge, this work proposes a new learning mechanism that leverages Vision-Language Models (VLMs) to empower UIE models with semantic-sensitive capabilities. To be concrete, our strategy first generates textual descriptions of key objects from a degraded image via VLMs. Subsequently, a text-image alignment model remaps these relevant descriptions back onto the image to produce a spatial semantic guidance map. This map then steers the UIE network through a dual-guidance mechanism, which combines cross-attention and an explicit alignment loss. This forces the network to focus its restorative power on semantic-sensitive regions during image reconstruction, rather than pursuing a globally uniform improvement, thereby ensuring the faithful restoration of key object features. Experiments confirm that when our strategy is applied to different UIE baselines, significantly boosts their performance on perceptual quality metrics as well as enhances their performance on detection and segmentation tasks, validating its effectiveness and adaptability.
\end{abstract}


\section{Introduction}

Underwater image processing is a critical technology in fields such as ocean exploration, biological monitoring, and underwater robotics \cite{gonzalez2023survey, wang2019experimental}. Due to severe degradation from light absorption and scattering, UIE has emerged as an essential pre-processing step. While recent deep learning-based methods have achieved remarkable success in producing visually pleasing results for human observers \cite{jiang2023perception}, a significant challenge has surfaced: a disconnect between perceptual image quality and the performance of machine cognition tasks. 
\begin{figure*}[t]
	\centering
	\includegraphics[width=2.1\columnwidth]{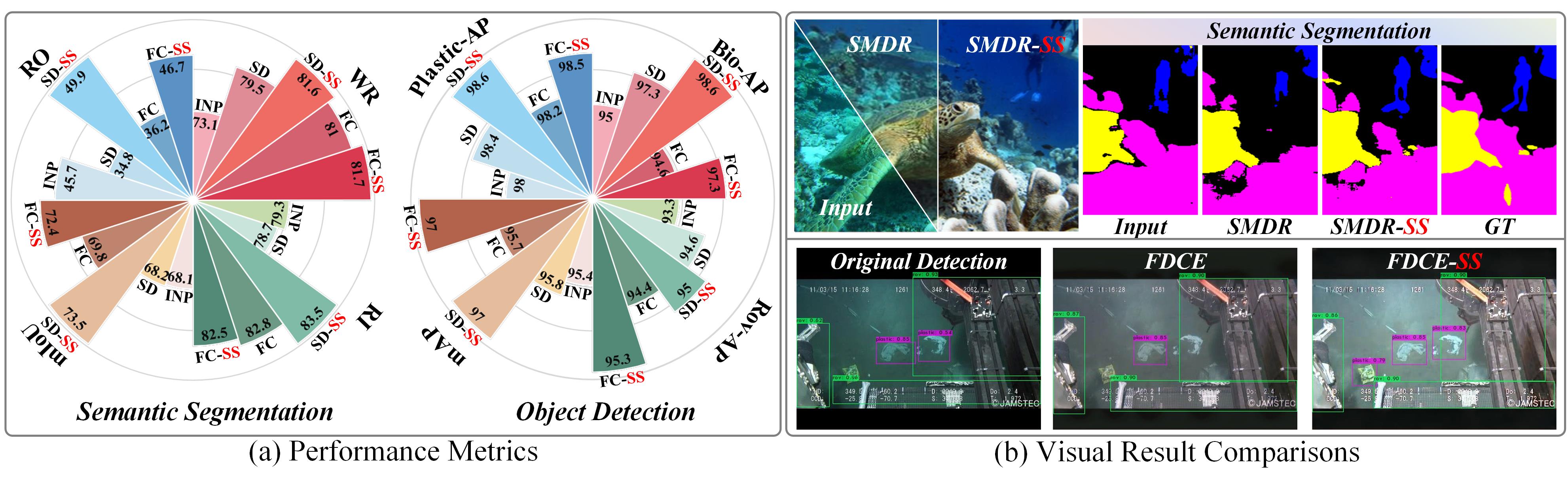}
	\caption{Impact of our semantic-sensitive strategy on downstream tasks. The radar charts on the left demonstrate consistent quantitative improvements in semantic segmentation and object detection when baseline models are empowered by our -SS strategy (in \textcolor{red}{Red}). Correspondingly, the qualitative examples on the right illustrate these benefits: our method leads to segmentation results closer to the Ground Truth (top) and enables more confident and accurate object detection (bottom).}
	\label{fig1}
\end{figure*}

We observe that state-of-the-art (SOTA) enhancement does not consistently translate to improved performance on tasks like object detection or segmentation \cite{tolie2024dicam, zhu2019adapting}, as illustrated in Fig. \ref{fig1}(b). This is due to the "task-agnostic" or "semantic-blind" nature of current UIE methods. In their pursuit of global, uniform enhancement, they often introduce imperceptible artifacts or cause a distribution shift that is misaligned with the data expectations of downstream models \cite{liu2019compounded, fu2022uncertainty}. This reveals their fundamental limitation: an inability to distinguish between semantic focal points (e.g., marine creatures, artifacts) and non-focal points (e.g., background water), leading to the corruption of semantic cues vital for machine understanding.

To solve this problem, researchers have begun to explore semantic-guided enhancement strategies. Early works leveraged semantic segmentation maps to guide feature learning or pixel fusion \cite{wu2023learning}. However, these methods depend heavily on high-quality, pixel-level annotations, which are notoriously scarce in underwater domains, risking the introduction of erroneous priors.

More recently, with the rise of VLMs, some have employed global, style-driven text prompts to guide image restoration \cite{pei2020consistency}. For instance, using a prompt like "a clear underwater photo" can improve overall image quality. While innovative, this approach is still essentially a "one-size-fits-all" strategy. It does not pay attention to the specific semantic content within the image, and thus cannot achieve the fine-grained, object-centric processing required for a truly robust enhancement. A clear gap remains in achieving targeted, content-aware enhancement.

To this end, we propose a powerful semantic-sensitive UIE strategy, the core of which is to use the open-world understanding ability of VLMs to empower the UIE model to perceive and focus on key semantic regions. Specifically, our mechanism first utilizes a VLM to generate textual descriptions of core objects from the input degraded image. Then, through a text-image alignment model, we accurately map these high-level semantic descriptions back to the 2D space of the image, thus generating a spatial semantic guidance map. Finally, this map is efficiently injected into the decoder of the UIE network through a dual-guidance mechanism, combining cross-attention and an explicit alignment loss. This guides the network to allocate more resources to semantically vital regions during image reconstruction. This design makes the enhancement process no longer a blind pursuit of global visual improvement, but an intelligent, content-aware restoration process that ensures the structure and texture of key objects are faithfully restored and enhanced. The main contributions of this paper can be summarized as follows:
\begin{itemize}
	\item We propose a VLM-driven, semantic-sensitive learning strategy to address the semantic blindness of traditional UIE, aiming to produce results that are robust for both human and machine perception.
	\item We design a dual-guidance mechanism to operationalize this strategy. The semantic map is used to simultaneously: structurally guide the network's information flow via a cross-attention injection module, and explicitly regularize its intermediate features via a new semantic alignment loss.
	\item Experimental validation demonstrates that our strategy not only improves perceptual quality but also boosts performance on subsequent machine cognition tasks, proving the effectiveness and adaptability of our strategy.
\end{itemize}

\section{Related Work}
In this section, we review prior image enhancement in three main areas: UIE techniques, downstream-aware image enhancement, and existing semantic-guided enhancement.
\subsection{Underwater Image Enhancement Techniques}
UIE aims to correct image degradation caused by absorption and scattering of water bodies. Early methods are mainly based on physical models. These methods recover clear images by inversely modeling the physical process of underwater imaging \cite{soni2020survey, wang2019experimental}. However, due to the diversity and complexity of the underwater environment, the physical prior assumptions of these methods are prone to lead to color distortion or noise amplification. In recent years, learning-based convolutional neural network (CNN) methods have become the mainstream in the field of UIE. These methods demonstrate powerful degradation feature extraction and image reconstruction capabilities by performing end-to-end learning on pairwise datasets \cite{zamir2020learning}. Learning methods have achieved remarkable success on traditional image quality metrics \cite{yang2023hifgan, ju2025towards}.  

\subsection{Image Enhancement for Downstream Tasks}
With the evolution of UIE technology, the academic community has increasingly recognized the discrepancy between enhancement goals centered on human vision and the requirements of downstream machine vision tasks \cite{liu2023image}. Directly applying SOTA UIE methods as a pre-processing step for object detection or segmentation tasks yields inconsistent performance improvements, sometimes even causing a decline \cite{zhou2023underwater,chen2020perceptual}. This "enhancement paradox" has spurred a new research direction: downstream-aware image enhancement. Research in this area is still in its nascent stages. Early explorations have primarily focused on joint training, where an enhancement network and a downstream task network are optimized simultaneously \cite{yu2023joint,kneubuehler2020flexible}. Although this approach enhances task relevance, it typically requires customizing the model for a specific downstream task, thus lacking flexibility and generality. Other works have attempted to design new loss functions aimed at preserving more features beneficial to downstream tasks during training \cite{liu2019underwater,huang2023underwater}. These efforts collectively point in a clear direction: for enhancement to be genuinely beneficial to downstream tasks, the process must transcend the pixel level and incorporate higher-dimensional information.

\subsection{Semantic-guided Image Enhancement}
Leveraging semantic information to guide image enhancement is becoming a popular strategy. Liao et al. \cite{liao2020guidance} used semantic segmentation maps to iteratively refine structural priors. Yan et al. \cite{yan2019semantic} utilized semantic segmentation to guide pixel fusion by assigning scaling factors to different regions. However, these methods rely heavily on high-quality, pixel-level annotated data to train the segmentation models. Such data is particularly scarce in underwater scenarios, which can lead to the introduction of erroneous semantic priors that misguide the enhancement process.

With the rise of VLMs, researchers have begun to exploit their powerful open-world understanding capabilities to guide image restoration \cite{zhou2025low,huang2024blenderalchemy, wang2023promptrestorer}. In the UIE domain, Liu et al. \cite{liu2024underwater} employed the CLIP model as a discriminator, using global text prompts to guide a generative model for image recovery. While this approach cleverly utilizes the cross-modal capabilities of VLMs, its guidance mechanism remains fundamentally global and style-driven. It does not concern itself with the specific objects present within the image and is therefore incapable of performing fine-grained, content-specific processing.

\section{Methodology}
\begin{figure}[t]
	\centering
	\includegraphics[width=1\columnwidth]{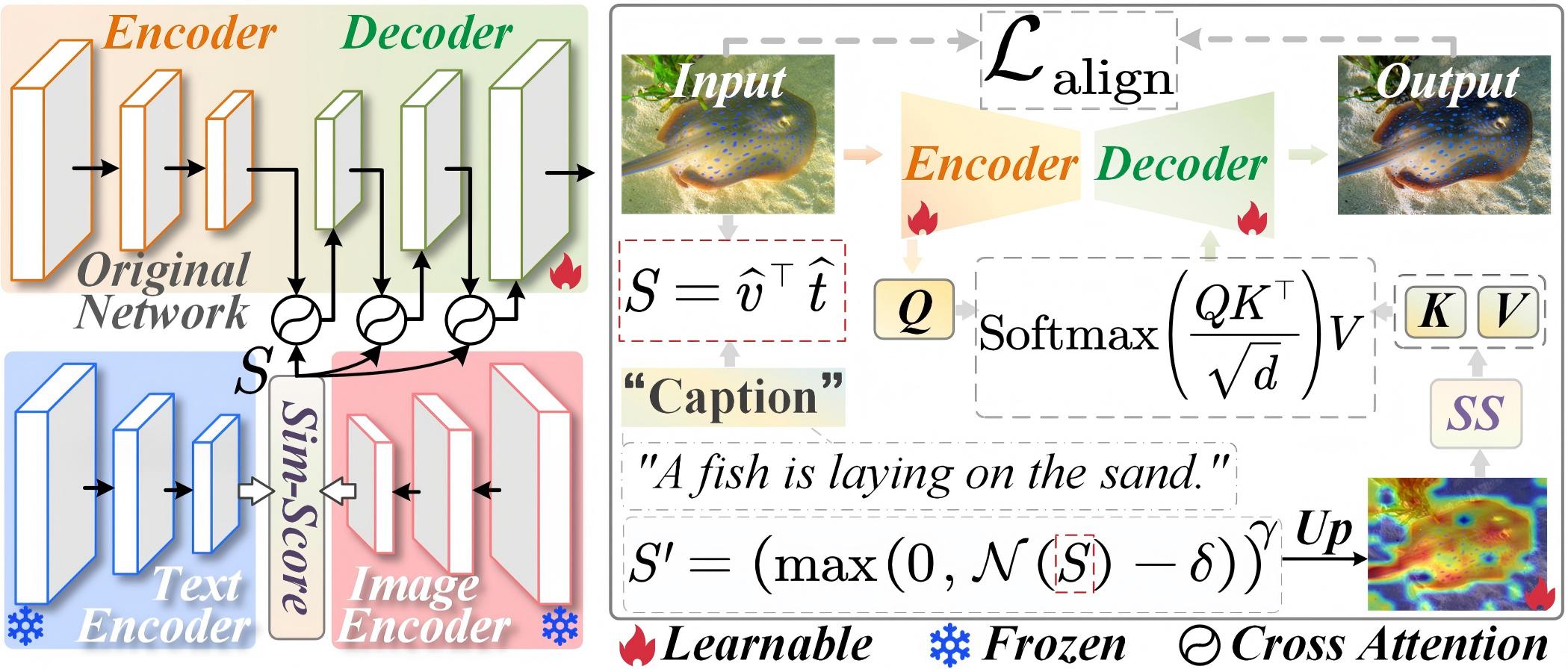}
	\caption{Overview of our semantic-sensitive learning strategy. A VLM-generated guidance map steers the UIE network via a dual-guidance approach: cross-attention and explicit supervision via an alignment loss.}
	\label{fig2}
\end{figure}

In this section, we introduce our strategy to empower UIE networks with semantic sensitivity, as shown in Fig. \ref{fig2}. We begin by defining the generation of a guidance map from VLM-derived text, creating an object-centric prior. We then elaborate on dual-guidance strategy: a cross-attention module provides architectural guidance, while a semantic alignment loss enforces feature-level supervision. 
\subsection{Generation of the Semantic Guidance Map}
To overcome the "task-agnostic" nature of UIE models, the network must first be made aware of what to focus on. We leverage VLMs to generate a reliable semantic prior that is specific to the content of each image, avoiding the need for scarce, densely annotated data required by traditional semantic segmentation.
We first generate a semantic guidance map, denoted as $M_{sem} \in \mathbb{R}^{H \times W}$, for each degraded underwater image $I_d \in \mathbb{R}^{H \times W \times 3}$. This map is designed to precisely quantify the semantic relevance between each spatial location in the image and the key objects of interest.
\subsubsection{Cross-Modal Feature Alignment}
Given the input image $I_d$ and its corresponding textual description $T$ generated by a VLM (adopting LLaVA \cite{liu2023visual}), we leverage the visual encoder $\Phi_v$ and text encoder $\Phi_t$ of a pre-trained vision-language alignment model (adopting BLIP \cite{li2022blip}) to extract features. $\Phi_v$ encodes $I_d$ into a set of patch features $F_v = \{f_v^1, f_v^2, \dots, f_v^N\} \in \mathbb{R}^{N \times C}$, while $\Phi_t$ encodes $T$ into a global textual feature vector $f_t \in \mathbb{R}^C$. After their respective projection layers and normalization, we obtain unit feature vectors $\hat{\mathbf{v}}_i$ and $\hat{\mathbf{t}}$.

\subsubsection{Semantic Similarity Calculation and Sharpening}
We compute the initial semantic relevance scores $s_i = \hat{\mathbf{v}}_i^\top \hat{\mathbf{t}}$ by taking the cosine similarity between each patch feature and the text feature. However, the raw similarity distribution can be too smooth to provide definitive guidance for the model \cite{zhao2024gradient}. To accentuate high-relevance regions and suppress irrelevant background, we design a semantic sharpening function, $\Psi_{\text{sharp}}$, which combines a power-law transformation with a thresholding operation:
\begin{equation}
	s'_i = \Psi_{\text{sharp}}(s_i; \gamma, \delta) = \left( \max(0, \mathcal{N}(s_i) - \delta) \right)^\gamma,
\end{equation}
where $\mathcal{N}(\cdot)$ denotes min-max normalization to the range $[0, 1]$, $\delta$ is a threshold to filter out low-relevance noise responses, and $\gamma > 1$ is the power-law coefficient used to non-linearly expand the score gap to focus attention. Finally, we reshape the 1D score sequence $S'=\{s'_1, \dots, s'_N\}$ and upsample it to the original image dimensions via interpolation, yielding the final single-channel semantic guidance map $M_{sem}$.
\subsection{Semantically-Guided Image Reconstruction}
With the guidance map generated, the next step is to integrate it into the UIE network. We propose a synergistic approach that combines structural guidance and explicit supervision to inject the semantic map into the decoder. Injecting semantic guidance here directly influences the decoding and reconstruction process, ensuring that as details are being rebuilt, priority is given to semantically important regions. This allows the semantic prior to be robustly and effectively utilized throughout the entire reconstruction process.
\begin{table*}[t]
	\renewcommand{\arraystretch}{1.2}
	
	\begin{center}
		{
			\setlength{\tabcolsep}{8.3pt} 
			\small
			\begin{tabular}{cccccccc} 
				
				\toprule[1.2pt]
				
				\multicolumn{1}{c}{\multirow{2}{*}[-0.8ex]{\Large{Methods}}} & \multicolumn{3}{c}{UIEB} & \multicolumn{2}{c}{U45} & \multicolumn{2}{c}{Challenge60}\\ 
				\cmidrule(r){2-4}  \cmidrule(r){5-6}  \cmidrule(r){7-8}
				~  & PSNR↑ & SSIM↑ & LPIPS↓& UIQM↑ & UCIQE↑ & UIQM↑ & UCIQE↑ \\ 
				\midrule[1.2pt]
				
				PUIE  & 21.05 & 0.869 &0.117 & 3.187 & 0.388 & 2.686 & 0.376 \\
				\textbf{PUIE-SS}  & 23.20{\tiny \textcolor{red}{(+2.15)}} & 0.884{\tiny \textcolor{red}{(+0.015)}} &0.092{\tiny \textcolor{red}{(-0.025)}} & 3.210{\tiny \textcolor{red}{(+0.023)}} & 0.392{\tiny \textcolor{red}{(+0.004)}} & 2.705{\tiny \textcolor{red}{(+0.019)}} & 0.379{\tiny \textcolor{red}{(+0.004)}} \\
				\midrule
				
				SMDR  & 22.44 & 0.899 &0.106 & 3.108 & 0.422 & 2.727 & 0.415 \\
				\textbf{SMDR-SS} & 23.28{\tiny \textcolor{red}{(+0.84)}} & 0.909{\tiny \textcolor{red}{(+0.010)}} & 0.099{\tiny \textcolor{red}{(-0.007)}} & 3.139{\tiny \textcolor{red}{(+0.031)}} & 0.436{\tiny \textcolor{red}{(+0.014)}} & 2.852{\tiny \textcolor{red}{(+0.125)}} & 0.433{\tiny \textcolor{red}{(+0.018)}} \\
				\midrule
				
				UIR  & 22.89 & 0.885 &0.124 & 2.874 & 0.374 & 2.458 & 0.371 \\
				\textbf{UIR-SS}  & \textbf{24.62}{\tiny \textcolor{red}{(+1.73)}} & 0.901{\tiny \textcolor{red}{(+0.016)}} &0.113{\tiny \textcolor{red}{(-0.011)}} & 3.016{\tiny \textcolor{red}{(+0.142)}} & 0.378{\tiny \textcolor{red}{(+0.004)}} & 2.659{\tiny \textcolor{red}{(+0.201)}} & 0.386{\tiny \textcolor{red}{(+0.015)}} \\
				\midrule
				
				PFormer  & 23.53 & 0.877 &0.113 & 3.105 & 0.437 & 2.426 & 0.426 \\
				\textbf{PFormer-SS} & 24.97{\tiny \textcolor{red}{(+1.44)}} & \textbf{0.933}{\tiny \textcolor{red}{(+0.056)}} & \textbf{0.087}{\tiny \textcolor{red}{(-0.026)}}& 3.126{\tiny \textcolor{red}{(+0.021)}} & \textbf{0.449}{\tiny \textcolor{red}{(+0.012)}} & 2.456{\tiny {\textcolor{red}{(+0.030)}}} & \textbf{0.437}{\tiny \textcolor{red}{(+0.011)}} \\
				\midrule
				
				FDCE  & 23.66 & 0.909 &0.111 & 3.258 & 0.434  & 3.015 & 0.418 \\
				\textbf{FDCE-SS} & 24.63{\tiny \textcolor{red}{(+0.97)}} & 0.927{\tiny \textcolor{red}{(+0.018)}} &0.093{\tiny \textcolor{red}{(-0.018)}} & \textbf{3.267}{\tiny \textcolor{red}{(+0.009)}} & 0.436{\tiny \textcolor{red}{(+0.002)}} & \textbf{3.079}{\tiny \textcolor{red}{(+0.064)}} & 0.421{\tiny \textcolor{red}{(+0.003)}} \\
				
				\bottomrule[1.2pt]
				
			\end{tabular}
		}
		\caption{Quantitative comparison of various UIE methods on the UIEB, U45, and Challenge60 datasets. -SS denotes empowering the baseline model with our proposed semantic-sensitive mechanism. Bold indicates the best performance among all methods, while \textcolor{red}{Red} values show the performance gain achieved by our mechanism.}
		\label{tab1}
	\end{center}
\end{table*}
\begin{figure*}[htbp]
	\centering
	\includegraphics[width=2\columnwidth]{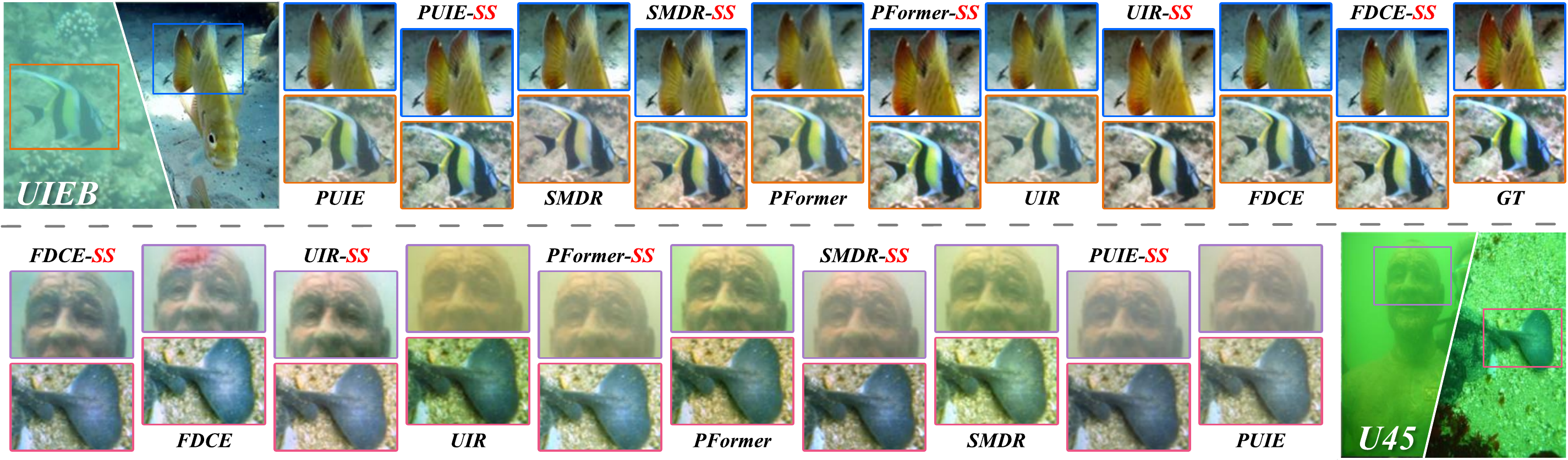}
	\caption{Visual comparison of enhancement results. Our -SS models produce images with better color fidelity and sharper details on key objects on the UIEB dataset (top). In challenging U45 scenes (bottom), our method effectively restores natural colors while avoiding artifacts introduced by baseline models. }
	\label{fig3}
\end{figure*}
\subsubsection{Cross Attention Injection Mechanism}
We inject semantic information into the UIE network's decoder via cross attention, allowing it to take effect at various decoding stages $l$. At stage $l$, the decoder feature $d_l$ acts as the query $Q_l$ by the linear projection to attend to the skip-connection feature $e_l$ from the encoder, which has been modulated by $M_{sem}$. Specifically, $M_{sem}$ is downsampled to the same spatial resolution as $e_l$, yielding $\tilde{M}^{(l)}$. This map is then used to element-wise weight $e_l$, which is then linearly projected to generate the key $K_l$ and value $V_l$. Based on this, the decoder can preferentially extract information from the semantically "illuminated" encoder features:
\begin{equation}
	d'_{l} = \text{softmax}\left(\frac{Q_l K_l^\top}{\sqrt{d_k}}\right)V_l.
\end{equation}

This design empowers the network with semantic-sensitivity and provides structural guidance.
\subsubsection{Explicit Semantic Alignment Loss}
While cross-attention provides structural guidance, its effect is implicit. To provide a more direct and measurable supervisory signal, we introduce an explicit semantic alignment loss, denoted as $\mathcal{L}_{\text{align}}$. It ensures the network is not just structurally guided but is also explicitly penalized for producing feature representations that deviate from the semantic prior. $\mathcal{L}_{\text{align}}$ operates directly on the intermediate feature maps of the decoder, forcing their spatial distribution to align with the semantic guidance map.

For the feature map $\mathbf{F}^{(l)}$ produced at stage $l$ of the decoder, our proposed $\mathcal{L}_{\text{align}}^{(l)}$ is composed of two terms:
\begin{equation}	
	\mathcal{L}_{\text{align}}^{(l)}(\mathbf{F}^{(l)}, \tilde{M}^{(l)}) = \underbrace{ \| \mathbf{F}^{(l)} \odot (1 - \tilde{M}^{(l)}) \|_F^2 }_{\text{Background}} - \underbrace{ \eta \langle \mathbf{F}^{(l)}, \tilde{M}^{(l)} \rangle }_{\text{Foreground}},
\end{equation}
where $\|\cdot\|_F^2$ is the squared Frobenius norm, $\langle \cdot, \cdot \rangle$ denotes the inner product, and $\eta$ is a hyperparameter balancing the two terms. For background suppression, this term penalizes the network for producing unnecessarily strong activations in non-key regions (defined by the mask $1 - \tilde{M}^{(l)}$) by minimizing the energy of the feature map $\mathbf{F}^{(l)}$, thereby suppressing background noise. For the foreground enhancement term, it rewards the network for producing strong responses in key object regions that are consistent with the semantic guidance by maximizing the correlation between $\mathbf{F}^{(l)}$ and $\tilde{M}^{(l)}$.
\begin{table*}[t]
	\renewcommand{\arraystretch}{1.2} 
	\begin{center}
		{ 
			
			\setlength{\tabcolsep}{3pt}
			
			\small 
			
			\begin{tabular}{ccccccccccc}
				
				\toprule[1.2pt]
				\multirow{2}{*}{\large{Methods}} &  \multicolumn{4}{c}{Object detection} & \multicolumn{6}{c}{Semantic Segmentation}\\ 
				\cmidrule(r){2-5}  \cmidrule(r){6-11}
				~ & Plastic-AP↑ & Bio-AP↑  & Rov-AP↑ & mAP↑ & RO↑ &RI↑ & FV↑ & WR↑ & HD↑& mIoU↑\\ 
				\midrule[1.2pt]
				
				\textit{Original} &98.00 &94.95 &93.33 &95.43 &45.68 &79.31 &66.44 &73.11 &75.97 &68.10 \\
				\hline
				
				PUIE  &97.90 &95.25 &93.07 &95.40 &33.31 &80.04 &68.61 &78.55  &70.49 &66.20  \\
				\textbf{PUIE-SS}  &97.84{\tiny \textcolor[HTML]{006400}{(-0.06)}} &97.13{\tiny \textcolor{red}{(+1.88)}} &93.86{\tiny \textcolor{red}{(+0.79)}} &96.28{\tiny \textcolor{red}{(+0.88)}} &38.84{\tiny \textcolor{red}{(+5.53)}} &82.36{\tiny \textcolor{red}{(+2.32)}} &75.33{\tiny \textcolor{red}{(+6.72)}} &81.31{\tiny \textcolor{red}{(+2.76)}} &76.15{\tiny \textcolor{red}{(+5.66)}} &70.80{\tiny \textcolor{red}{(+4.60)}}\\
				\hline 
				
				UIR  &96.49 &92.43 &94.22 &94.37 &34.54 &82.97 &74.78 &79.19 &71.08 &68.52 \\
				\textbf{UIR-SS}  &97.22{\tiny \textcolor{red}{(+0.73)}} &93.96{\tiny \textcolor{red}{(+1.53)}} &94.77{\tiny \textcolor{red}{(+0.55)}} &95.31{\tiny \textcolor{red}{(+0.94)}} &42.03{\tiny \textcolor{red}{(+7.49)}} &\textbf{84.78}{\tiny \textcolor{red}{(+1.81)}} &74.91{\tiny \textcolor{red}{(+0.12)}} &79.06{\tiny \textcolor[HTML]{006400}{(-0.13)}} &71.49{\tiny \textcolor{red}{(+0.41)}} &70.45{\tiny \textcolor{red}{(+1.93)}}\\
				\hline 
				
				PFormer  &98.20 &96.29 &94.97 &95.50 &36.23 &83.83 &75.58 &80.90 &70.12 &69.34 \\
				\textbf{PFormer-SS}  &98.48{\tiny \textcolor{red}{(+0.28)}} &96.99{\tiny \textcolor{red}{(+0.70)}} &95.15{\tiny \textcolor{red}{(+0.18)}} &96.87{\tiny \textcolor{red}{(+1.37)}} &\textbf{51.52}{\tiny \textcolor{red}{(+15.29)}} &83.34{\tiny \textcolor[HTML]{006400}{(-0.49)}} &\textbf{77.64}{\tiny \textcolor{red}{(+2.06)}} &\textbf{83.43}{\tiny \textcolor{red}{(+2.53)}} &\textbf{77.80}{\tiny \textcolor{red}{(+7.67)}} &\textbf{74.75}{\tiny \textcolor{red}{(+5.41)}}\\
				\hline
				
				SMDR  &98.39 &97.25 &94.61 &95.76 &34.78 &78.66 &73.08 &79.47 &74.92 &68.18 \\
				\textbf{SMDR-SS}  &\textbf{98.59}{\tiny \textcolor{red}{(+0.20)}} &\textbf{98.59}{\tiny \textcolor{red}{(+1.34)}} &95.01{\tiny \textcolor{red}{(+0.40)}} &96.98{\tiny \textcolor{red}{(+1.22)}} &49.94{\tiny \textcolor{red}{(+15.17)}} &83.48{\tiny \textcolor{red}{(+4.82)}} &76.99{\tiny \textcolor{red}{(+3.91)}} &81.55{\tiny \textcolor{red}{(+2.09)}} &75.58{\tiny \textcolor{red}{(+0.67)}} &73.51{\tiny \textcolor{red}{(+5.33)}}\\
				\hline 
				
				FDCE  &98.17 &94.55 &94.44 &95.72 &36.24 &82.81 &76.11 &80.96 &72.80 &69.78 \\
				\textbf{FDCE-SS}  &98.50{\tiny \textcolor{red}{(+0.33)}} &97.27{\tiny \textcolor{red}{(+2.72)}} &\textbf{95.27}{\tiny \textcolor{red}{(+0.83)}} &\textbf{97.01}{\tiny \textcolor{red}{(+1.29)}} &46.71{\tiny \textcolor{red}{(+10.47)}} &82.54{\tiny \textcolor[HTML]{006400}{(-0.27)}} &75.65{\tiny \textcolor[HTML]{006400}{(-0.46)}} &81.73{\tiny \textcolor{red}{(+0.78)}} &75.17{\tiny \textcolor{red}{(+2.37)}} &72.36{\tiny \textcolor{red}{(+2.58)}}\\
				
				\bottomrule[1.2pt]
			\end{tabular}
		}
		\caption{Quantitative evaluation on downstream tasks, comparing UIE baselines with and without our -SS strategy. The tasks include object detection, and semantic segmentation.  \textcolor{red}{Red} denotes improvement over the baseline, while \textcolor[HTML]{006400}{Green} indicates a decline.} 
		\label{tab2}
	\end{center}
\end{table*}
\begin{figure*}[t]
	\centering
	\includegraphics[width=2.1\columnwidth]{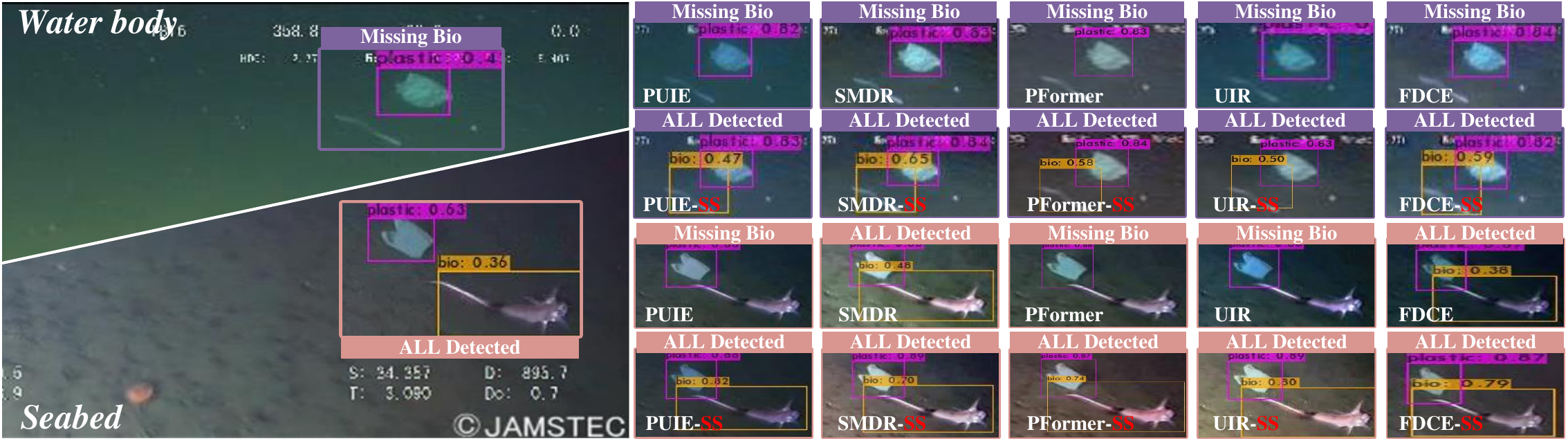}
	\caption{Visual comparison on the object detection task. Our -SS enhancement significantly improves the detection of small, low-contrast objects in both murky water body (top) and complex seabed (bottom) environments, effectively mitigating the missed detection issue prevalent in baseline methods. }
	\label{fig4}
\end{figure*}
\subsection{Overall Training Objective}
To produce a final result that is both visually pleasing and semantically sound, the training objective must balance low-level pixel fidelity, high-level perceptual quality, and our proposed semantic alignment.

The final objective function, $\mathcal{L}_{\text{total}}$, is a weighted sum of a reconstruction loss, $\mathcal{L}_{\text{recon}}$, and our proposed semantic alignment loss, $\mathcal{L}_{\text{align}}$:
\begin{equation}
	\mathcal{L}_{\text{total}} = \mathcal{L}_{\text{recon}} + \lambda_{\text{align}} \sum_{l \in L} \mathcal{L}_{\text{align}}^{(l)},
\end{equation}
where $l$ represents the set of decoder stages where the semantic alignment loss is applied, and $\lambda_{\text{align}}$ is a key hyperparameter that balances the two loss terms and we empirically set $\lambda_{\text{align}}$ to 0.1.

The reconstruction loss, $\mathcal{L}_{\text{recon}}$, is itself a composite of a L1 loss and a perceptual loss, $\mathcal{L}_{\text{percep}}$. While the L1 loss ensures basic pixel fidelity, the perceptual loss guarantees structural and textural similarity by comparing the enhanced and reference images in the feature space of the pre-trained VGG-19 network \cite{simonyan2014very}:
\begin{equation}
	\mathcal{L}_{\text{recon}} = \| I_e - I_{gt} \|_1 + \lambda_{\text{percep}} \sum_{j} \| \phi_j(I_e) - \phi_j(I_{gt}) \|_1,
\end{equation}
where $\phi_j$ denotes the feature extractor from the $j$-th convolutional layer of the VGG-19, and $\lambda_{\text{percep}}$ is its weight. By optimizing this objective, the empowered network learns to restore accurate pixel values while preserving perceptual features and ensuring its internal representations align with the semantic prior. This holistic training strategy results in an enhancement that is visually superior and more robust for both human and machine perception.
\begin{figure*}[t]
	\centering
	\includegraphics[width=2.1\columnwidth]{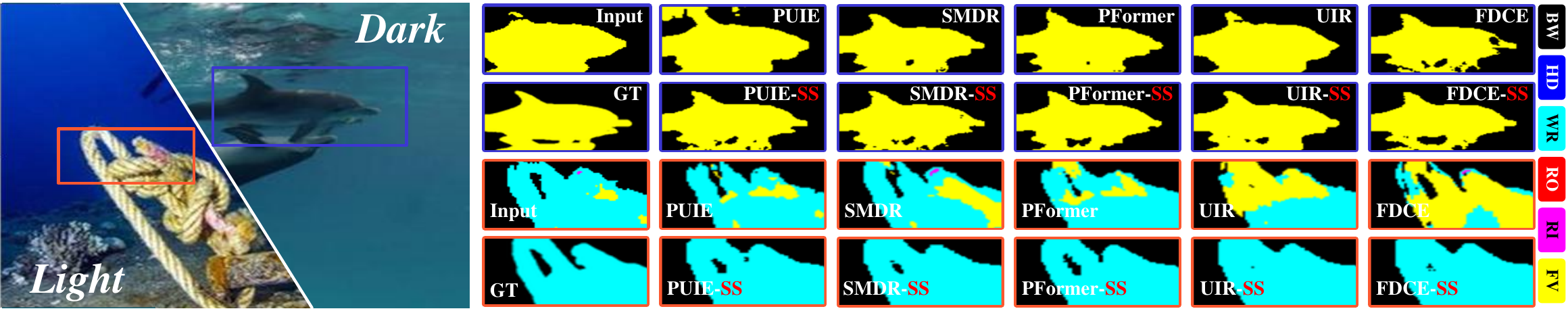}
	\caption{Visual comparison on the semantic segmentation task. Our semantic-sensitive enhancement preserves object boundaries and reduces background confusion in both dark (top row) and light (bottom row) scenes, leading to more accurate segmentation masks compared to baseline methods.}
	\label{fig5}
\end{figure*}
\section{Experiments}
In this section, we conduct comprehensive experiments to validate the effectiveness of our strategy. We present our experimental setup (datasets, metrics, implementation), followed by extensive quantitative and qualitative comparisons on both perceptual quality and downstream vision tasks. Finally, ablation studies analyze the contributions of proposed strategy.
\subsection{Datasets and Evaluation Metrics}
\subsubsection{Datasets for UIE Tasks.}To train and evaluate our models, we utilize a combination of standard UIE benchmarks. We use the UIEB dataset \cite{li2019underwater} for supervised training, leveraging its training set of 790 paired images. For full-reference evaluation, we test on its official test set of 100 paired images. To assess the generalization capability and enhancement quality in real-world scenarios without ground-truth, we test the models trained on UIEB on two additional no-reference datasets: U45 \cite{li2019fusion} and Challenge60 \cite{li2019underwater}.
\subsubsection{Datasets for Downstream Vision Tasks.}To validate that our enhancement produces semantically robust images that are more conducive to machine understanding, we evaluate its impact on two distinct downstream tasks using their benchmark datasets: Trash-ICRA19 dataset \cite{fulton2019robotic} for object detection, SUIM dataset \cite{islam2020semantic} for semantic segmentation.
\subsubsection{Evaluation Metrics.}Our evaluation protocol is comprehensive, covering both perceptual quality and downstream task performance. For UIE tasks, we measure full-reference quality on the UIEB test set using the peak signal-to-noise ratio (PSNR), the structural similarity index measure (SSIM) and the learned perceptual image patch similarity (LPIPS) \cite{zhang2018unreasonable}. For no-reference evaluation on U45 and Challenge60, we employ the underwater image quality measure (UIQM) \cite{panetta2015human} and the underwater color image quality evaluation (UCIQE) \cite{yang2015underwater}. 
For downstream vision tasks, we report the average precision (AP) for three distinct categories (Plastic, Bio, Rov) and the overall mean average precision (mAP) for object detection; we report the intersection over union (IoU) for five categories: Robots (RO), Reefs and invertebrate (RI), Fish and vertebrates (FV), Wrecks or ruins (WR), Human divers (HD) and the overall mean intersection over union (mIoU) for semantic segmentation.

\subsection{Implementation Details}
Our semantic-sensitive mechanism is designed to be a versatile and pluggable module that can be integrated into various UIE architectures. To demonstrate its broad applicability, we select five recent, SOTA UIE models that feature an encoder-decoder structure as baselines: PUIE \cite{fu2022uncertainty}, SMDR \cite{zhang2024synergistic}, UIR \cite{guo2025underwater}, PFormer \cite{khan2025phaseformer}, and FDCE \cite{cheng2024fdce}.
\subsubsection{Training for UIE Task.}For our primary enhancement experiments, we first train these five baseline models on the UIEB training set to establish their original performance and obtain their official weights. Subsequently, we integrate our semantic-sensitive mechanism into each of these models. We then retrain these empowered models, denoted with the suffix -SS (for Semantic-Sensitive), on the same UIEB dataset. All original and -SS models are then evaluated on the test sets of UIEB, U45, and Challenge60.
\subsubsection{Evaluation on Downstream Tasks.}For downstream task evaluation, we use pre-trained weights of both the original baseline models and our -SS variants as image pre-processors. For each task, the input images from the respective test set are first enhanced by five original and five -SS models. The resulting enhanced images are then fed into a unified downstream network for evaluation. This rigid setup ensures that any observed performance difference is strictly attributable to the intrinsic quality and semantic integrity of the enhancement provided by the low-level model.

\subsection{Quantitative Evaluation}
\subsubsection{Performance on UIE Tasks.}
Table \ref{tab1} presents the quantitative results on the UIE datasets. For UIEB, our mechanism shows improvements in both PSNR and SSIM across all five baselines. This indicates that by focusing on semantically important regions, our method can achieve higher fidelity in image reconstruction. For U45 and Challenge60, our method also performs excellently on the UIQM and UCIQE metrics. Although there are fluctuations, the overall trend is positive. This demonstrates that the semantic-guided approach does not overfit to a specific data distribution and can maintain better color balance and clarity in challenging scenes.
\subsubsection{Performance on Downstream Tasks.}
Table \ref{tab2} summarizes the evaluation results on downstream vision tasks. The consistent outperformance of our empowered versions serves as strong evidence that our strategy produces images with more robust features for machine analysis. Notably, while some baselines struggle to outperform the Original unenhanced images, our method consistently delivers significant gains over both. The improvement is particularly pronounced in semantic segmentation, where our method provides cleaner object structures for pixel-level classification, and also consistent in object detection, making targets more identifiable.
\subsection{Qualitative Analysis}
\subsubsection{Comparisons on UIE Tasks.}Fig. \ref{fig3} presents a visual comparison of the enhancement results on the UIEB and U45 datasets. For UIEB, while all baseline methods improve the overall brightness and color, they exhibit suboptimal performance in preserving the fine details of the key object (the fish). PUIE and UIR tend to result in over-saturated colors, whereas SMDR and PFormer sometimes flatten the textures.
For U45, the original images suffer from severe color cast and low visibility. The baseline methods either fail to sufficiently remove the greenish color cast or introduce unnatural artifacts. In contrast, our empowered models focus their restoration capabilities on semantically salient regions. Consequently, they exhibit sharper details, more natural color transitions, and a clearer separation from the background, consistently producing results that are more visually pleasing.
confirms that our enhancement strategy makes it easier for downstream models to identify and locate key targets.
\begin{figure}[t]
	\centering
	\includegraphics[width=1\columnwidth]{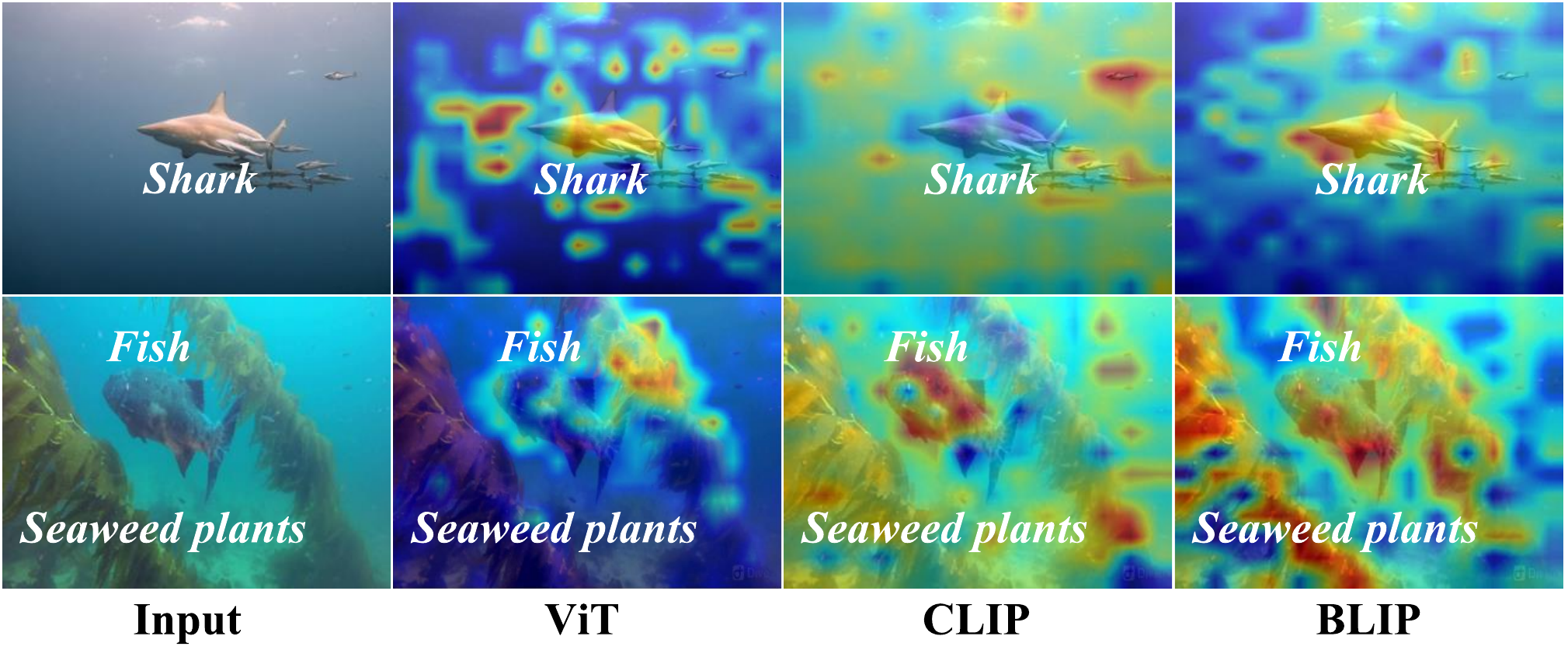}
	\caption{Visual comparison of guidance maps from different models.}
	\label{fig6}
\end{figure}

\subsubsection{Comparisons on Downstream Tasks.}Fig. \ref{fig4} and Fig. \ref{fig5} visually demonstrate the performance improvements on various downstream tasks.
Fig. \ref{fig4} presents a marine debris detection scenario featuring small, low-contrast objects against a complex seabed background. When using the original baseline enhancer, the detection model frequently fails. However, after applying our semantic-sensitive mechanism, the same object is successfully detected in all cases. This indicates that the empowered model provides targeted local contrast enhancement that is sufficient to identify previously missed objects.
Fig. \ref{fig5} illustrates the impact on semantic segmentation. In the dark scene, both PUIE-SS and SMDR-SS models yield superior segmentation results by effectively enhancing the contrast between the dolphin and the dark body of water. Conversely, in the bright scene, baseline methods tend to over-enhance the entire scene, which leads to mis-segmentation errors. In contrast, our empowered models, by selectively enhancing the target, maintain a clear distinction between the foreground object and the background.

\subsection{Ablation Study}

\subsubsection{Analysis of Text-Image Alignment for Semantic Guidance.} To justify our choice of model for generating the semantic guidance map, a critical precursor in our strategy. As shown in Fig. \ref{fig6}, we compare guidance maps generated by three different vision-language alignment approaches: a standard Vision Transformer (ViT) using class attention, CLIP, and BLIP. The class-based attention of the ViT produces coarse and diffuse heatmaps, failing to precisely localize objects or disentangle concepts like "fish" and "seaweed." While contrastive learning endows CLIP with sharper attention, it also induces noise-prone spurious activations in background regions. In stark contrast, BLIP's fusion-based alignment strategy yields superior results. It generates clean, sharply-defined, and spatially accurate maps that correctly highlight the objects described in the text prompt with minimal noise. This high-fidelity spatial guidance is crucial for the effectiveness of our subsequent enhancement stages, validating our selection of BLIP.
\subsubsection{Analysis of Semantic Guidance Injection Stage.}
\begin{figure}[t]
	\centering
	\includegraphics[width=1\columnwidth]{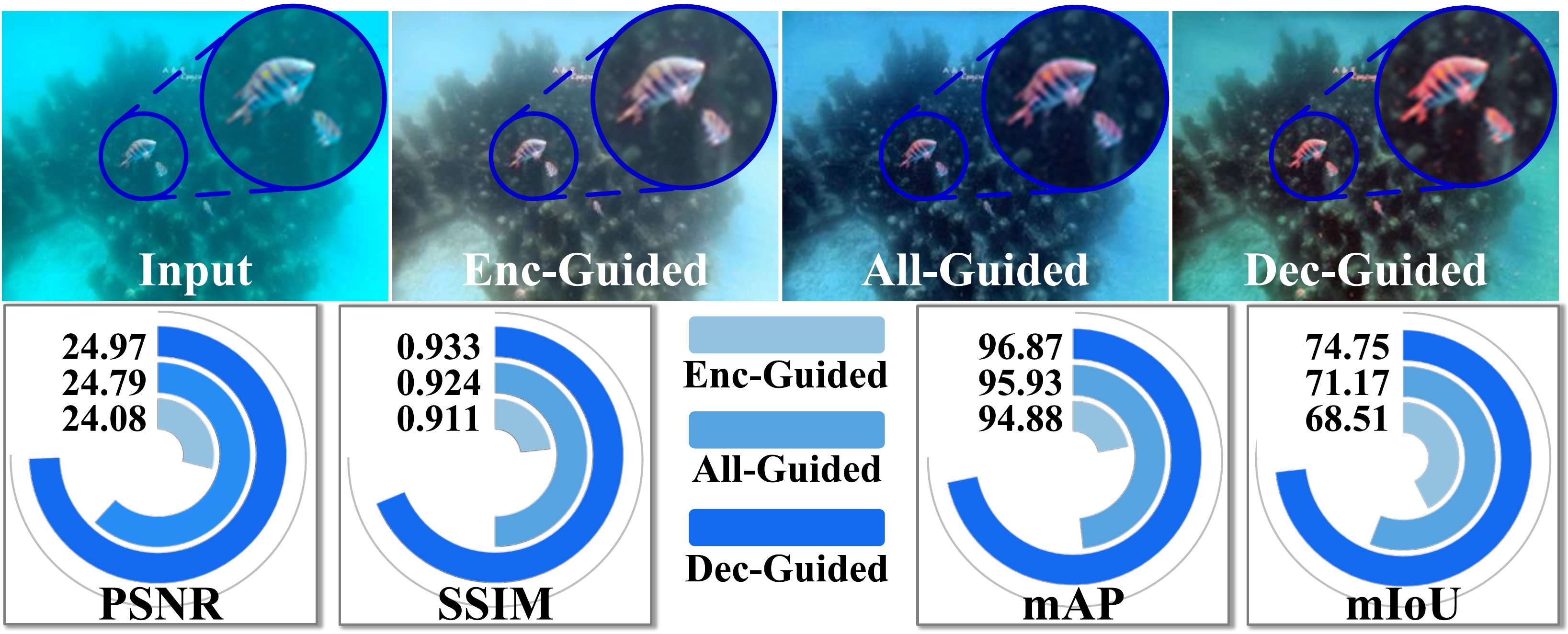}
	\caption{Visual comparison of the semantic guidance injection stage.}
	\label{fig7}
\end{figure}

To validate our strategy of injecting guidance into the decoder, we ablate three configurations: injecting into the Encoder only, All stages, or the Decoder only. As shown in Fig. \ref{fig7}, our decoder-only approach is demonstrably superior. Qualitatively, it produces the most visually pleasing result with vibrant colors and sharp object details. Quantitatively, it achieves the highest scores across all metrics, including perceptual quality and machine cognition. This confirms our hypothesis that injecting semantic guidance during the decoder's reconstruction process is the most effective strategy, as it directly steers image formation, whereas injecting it during the encoder's feature extraction stage is less effective and can even be counter-productive.
\section{Conclusion}
This paper addresses the semantic blindness of traditional UIE methods, a core limitation that hinders their performance in both human perception and machine cognition. We propose a novel semantic-sensitive learning strategy that empowers UIE models with content-aware capabilities. Our strategy first leverages a VLM to generate a spatial semantic guidance map. It then steers the enhancement process through a dual-guidance mechanism, which synergistically combines a cross-attention module for structural guidance and an explicit semantic alignment loss for feature-level supervision. Extensive experiments demonstrate that our strategy is both effective and adaptable. When applied to various UIE baselines, it produces visually superior results and significantly boosts performance on subsequent detection and segmentation tasks. 

\section{Acknowledgments}
This work is supported by the National Natural Science Foundation of China under grant Nos. U24A20219, 62272281, 62576193, 62202268, and 62301105; special funds of Taishan Scholars Project of Shandong Province under grant Nos. tsqn202306274 and tsqn202507240; Natural Science Foundation of Shandong Province under grant Nos. ZR2025QC712, ZR2025QC695 and ZR2025MS985.

\bibliography{aaai2026}

\end{document}